%% file: Main.tex
\documentclass[conference]{IEEEtran}
\IEEEoverridecommandlockouts

\usepackage{cite}
\usepackage{amsmath,amssymb,amsfonts}
\usepackage{algorithmic}
\usepackage{graphicx}
\usepackage{textcomp}
\usepackage{xcolor}
\usepackage{subfiles}
\usepackage{booktabs}
\usepackage{multirow}
\usepackage{booktabs}
\usepackage{multirow}
\usepackage{pifont}
\usepackage[normalem]{ulem}
\usepackage{colortbl}
\usepackage{longtable}
\usepackage{subfiles}

\def\BibTeX{{\rm B\kern-.05em{\sc i\kern-.025em b}\kern-.08em
    T\kern-.1667em\lower.7ex\hbox{E}\kern-.125emX}}
\begin{document}

\title{A Dual-Perspective Metaphor Detection Framework Using Large Language Models}

\author{\IEEEauthorblockN{1\textsuperscript{st} Yujie Lin$^\star$}
\IEEEauthorblockA{\textit{School of Informatics
} \\
\textit{Xiamen University}\\
Xiamen, China \\
yjlin@stu.xmu.edu.cn}
\and
\IEEEauthorblockN{2\textsuperscript{nd} Jingyao Liu$^\star$}
\IEEEauthorblockA{\textit{School of Informatics
} \\
\textit{Xiamen University}\\
Xiamen, China \\
liujingyao@stu.xmu.edu.cn}
\and
\IEEEauthorblockN{3\textsuperscript{rd} Yan Gao}
\IEEEauthorblockA{\textit{School of Informatics
} \\
\textit{Xiamen University}\\
Xiamen, China \\
gaoyan@stu.xmu.edu.cn}
\and
\IEEEauthorblockN{4\textsuperscript{th} Ante Wang}
\IEEEauthorblockA{\textit{School of Informatics} \\
\textit{Xiamen University}\\
Xiamen, China \\
wangante@stu.xmu.edu.cn}
\and
\IEEEauthorblockN{5\textsuperscript{th} Jinsong Su$^\dag$ }
\IEEEauthorblockA{\textit{School of Informatics} \\
\textit{Xiamen University}\\
Xiamen, China \\
jssu@xmu.edu.cn}
\thanks{$\star$ Equal Contribution.

$\dag$ Corresponding Author.

The project was supported by 
National Natural Science Foundation of China (No. 62276219),  
Natural Science Foundation of Fujian Province of China (No. 2024J011001),
and
the Public Technology Service Platform Project of Xiamen (No.3502Z20231043).
We also thank the reviewers for their insightful comments.}
}

\maketitle

\begin{abstract}
Metaphor detection, a critical task in natural language processing, involves identifying whether a particular word in a sentence is used metaphorically. Traditional approaches often rely on supervised learning models that implicitly encode semantic relationships based on metaphor theories. However, these methods often suffer from a lack of transparency in their decision-making processes, which undermines the reliability of their predictions. Recent research indicates that LLMs (large language models) exhibit significant potential in metaphor detection. Nevertheless, their reasoning capabilities are constrained by predefined knowledge graphs. To overcome these limitations, we propose DMD, a novel 
dual-perspective framework that harnesses both implicit and explicit applications of metaphor theories to guide LLMs in metaphor detection and adopts a self-judgment mechanism to validate the responses from the aforementioned forms of guidance. In comparison to previous methods, our framework offers more transparent reasoning processes and delivers more reliable predictions. Experimental results prove the effectiveness of DMD, demonstrating state-of-the-art performance across widely-used datasets.
\end{abstract}


\section{Introduction}

A metaphor is a widespread rhetorical device where one concept is described by referring to another, thereby creating a figurative comparison~\cite{lakoff2008metaphors,lagerwerf2008openness}. Metaphors are prevalent in everyday communication and are often used to vividly express emotions and viewpoints. Therefore, metaphor detection, which aims to automatically determine whether a specific word in a given sentence is used metaphorically or literally, has become a crucial task in the field of natural language processing.

To precisely detect metaphors, researchers have proposed several metaphor theories, among which \textit{Metaphor Identification Procedure~(MIP)}~\cite{group2007mip,mip2010,mip2016} and \textit{Selectional Preference Violation~(SPV)}~\cite{spv1,spv2} theories are the most critical. The MIP theory posits that a metaphor arises when the target word's basic meaning conflicts with its contextual meaning. In contrast, the SPV theory determines metaphors based on the statistical anomalies in word combinations. By analyzing large corpora, common co-occurrence patterns of words establish the selectional preference. When the context deviates significantly from the selectional preference, it is regarded as a potential metaphor.

Grounded in these metaphor theories, traditional metaphor detection methods predominantly employ supervised learning to implicitly model the semantic relationships between a target word and its context. For example, MelBERT~\cite{melbert-tra3} derives two feature vectors aligned with MIP and SPV theories for metaphor detection. Building on this, MisNet~\cite{misnet} and MiceCL~\cite{tra4} enhance the modeling of the two feature vectors, respectively, contributing to further advancements in metaphor detection. Although these supervised learning methods have achieved promising results, they merely provide prediction outcomes without transparency in the decision-making process, thereby undermining the credibility of their predictions. With the development of large language models~(LLMs), recent studies~\cite{cot,self-consis,self-ref,plan_and,least2most, ref11, ref12} have demonstrated the remarkable reasoning abilities of LLMs across a wide range of tasks~\cite{gpt4,chatgpt,llama}. In this context, Tian et al.~\cite{TSI} pioneered the application of LLMs for metaphor detection by predefining a metaphor theory knowledge graph and designing a series of guiding questions based on this graph. Although their work showed the potential of LLMs in metaphor detection, it was constrained by the reliance on a predefined knowledge graph, which limited the comprehensive utilization of LLMs' strong reasoning capabilities.

\subfile{imgs/main_fig}

To address the aforementioned issues, we propose a novel \textbf{D}ual-Perspective \textbf{M}etaphor \textbf{D}etection~\textbf{(DMD)} framework that leverages LLMs from both implicit and explicit perspectives to provide more reliable answers based on metaphor theories including MIP and SPV. As shown in Figure \ref{fig:main_fig}, our proposed DMD comprises three parts: Implicit Theory-Driven Guidance, Explicit Theory-Driven Guidance, and Self-Judgment. In Implicit Theory-Driven Guidance, we first construct a datastore consisting of a collection of key-value pairs. The value is a sample from an annotated dataset for metaphor detection, and the key is a high-dimensional vector representing the features learned from metaphor theories, computed by a pre-trained metaphor detection mode, such as MelBERT. During inference, for each input sample, we compute the feature vector as mentioned above, then search the built datastore for the top-$k$ samples with similar features. These retrieved samples are then used as reference samples, along with the inference sample, to form a prompt for the LLM, yielding an answer and its corresponding explanation. By doing so, the LLM leverages the implicit theoretical similarities drawn from the retrieved samples to make more informed decisions. In Explicit Theory-Driven Guidance, we begin by retrieving the definition and relevant usage examples of the target word’s lemma , and explicitly provide the LLM with the contents of metaphor theories to generate multi-step thoughts. By combining the multi-step thoughts and the retrieved information into a structured prompt, the LLM is able to produce a more detailed and informative response. To further synthesize the predictions from these two perspectives, we finally design Self-Judgment, where the LLM acts as a judge to evaluate the answers and explanations provided by both forms of guidance, thereby ensuring the delivery of a coherent and substantiated answer with a comprehensive explanation.

Compared with previous studies, our framework is the first to leverage both implicit and explicit perspectives to jointly utilize metaphor theories for guiding LLMs on metaphor detection. It addresses the absence of decision-making processes in traditional supervised learning methods and the reliance on fixed rules in prior LLM-based methods that restrict the full utilization of LLMs' reasoning capabilities. Furthermore, our framework exploits the advanced semantic understanding abilities of LLMs to validate the responses from both perspectives, thus enhancing the credibility of the predictions. We conduct experiments on the widely-used MOH-X and TroFi datasets, where our framework demonstrated outstanding performance, achieving state-of-the-art~(SOTA) results.

\section{Methods}

\subsection{Overview}

In this section, we will introduce the details of our DMD framework. As depicted in Figure \ref{fig:main_fig}, there are three parts: \textit{Implicit Theory-Driven Guidance}, \textit{Explicit Theory-Driven Guidance} and \textit{Self-Judgment}. Initially, we employ the implicit and explicit theory-driven guidance to generate answers and explanations rooted in metaphor theories, which are subsequently validated through the self-judgment mechanism, resulting in a more trustworthy final answer.

\subsection{Implicit Theory-Driven Guidance}

As shown in Figure \ref{fig:main_fig}(a), in view of the strong encoding capabilities of supervised learning models~\cite{bert,liu2019roberta,jia2024md}, we utilize a pre-trained model to extract the representations learned from MIP and SPV theories for each sample within a large annotated dataset. These representations, along with their corresponding samples, are stored in a datastore. During inference, we retrieve the top-$k$ most similar representations from the datastore and adopt the corresponding $k$ samples for in-context learning~\cite{icl}. In this manner, the LLM effectively uncovers the implicit similarities based on the theoretical relationships between the inference sample and the retrieved samples to obtain more accurate predictions.

\subsubsection{Datastore Creation}

Our datastore is built offline and comprises a collection of key-value pairs. Each value is an entire sample from VUA18~\cite{vua18}, a comprehensive annotated dataset for metaphor detection, and the paired key is a high-dimensional representation learned from the MIP and SPV theories, computed by a pre-trained MelBERT~\cite{melbert-tra3} model. 

Formally, in the annotated dataset $\mathcal{D}$, each sample $e$ consists of a sentence $s=\{w_1,w_2,...,w_n\}$ composed of $n$ words, a target word $w_t$, and a corresponding label $y$. Firstly, we employ the encoder of MelBERT to capture a set of contextualized embedding vectors $\{v_s, v_{s,1}, ..., v_{s,t}, ..., v_{s,n} \}$ by inputting $s$, and the target word embedding vector $v_t$ by inputting only $w_t$: 

\begin{equation}
v_s, v_{s,1},\ldots , v_{s,t}, \ldots, v_{s,n}  = \mathcal{E}(s) ,
\label{eq:MelBERT-0}
\vspace{-5pt}
\end{equation}

\begin{equation}
v_t = \mathcal{E}(w_t) ,
\label{eq:MelBERT-1}
\end{equation}
where $\mathcal{E}$ is the encoder of MelBERT. Notably, $v_s$ donates the global representation of the entire sentence.

Then, we use MIP theory to learn the representation $h_{MIP}$ by concatenating $v_{s,t}$ and $v_t$. Similarly, SPV theory is employed to learn the representation $h_{SPV}$ by concatenating $v_s$ and $v_{s,t}$:

\begin{equation}
h_{MIP} = f([v_{s,t}; v_t]) ,
\label{eq:MelBERT-2}
\vspace{-5pt}
\end{equation}

\begin{equation}
h_{SPV} = g([v_s; v_{s,t}]) ,
\label{eq:MelBERT-3}
\end{equation}
where $f(\cdot)$ is an MLP layer modelling the gap between the vectors $v_{s,t}$ and $v_t$, while $g(\cdot)$ represents another MLP layer capturing the semantic difference between vectors $v_s$ and $v_{s,t}$.

These representations are finally concatenated to form the complete representation $h_T$:

\begin{equation}
h_{T} = [h_{MIP}; h_{SPV}] .
\label{eq:MelBERT-4}
\end{equation}

Upon completing the above steps, the representation $h_T$ serves as the key and the entire sample $e$ serves as the value. The complete datastore is formally defined as follows:
\begin{equation}
(\mathcal{K}, \mathcal{V}) = \{(h_T, e) | e \in \mathcal{D}\} .
\label{eq:Datastore}
\end{equation}

\subsubsection{Inference}

During the inference phase, given a sample consisting of a sentence $s$ and a target word $w_t \in s$, the pre-trained MelBERT produces the representation $h_T$ as previously described. This representation is then used to query the pre-established datastore $\mathcal{D}$ to find $k$ nearest neighbors $\mathcal{N}=\{e_1, ..., e_k\}$ according to squared-$L^2$ distance. 

Once $k$ samples are retrieved, they are combined with the inference sample and provided to the LLM to generate the response $R_{im}$ including an answer and its corresponding explanation:

\begin{equation}
R_{im}=\mathcal{F}(ins_{im}, \mathcal{N}, s, w_t) ,
\label{eq:ICL}
\end{equation}
where $\mathcal{F}$ represents the LLM and $ins_{im}$ means the instruction for Implicit Theory-Driven Guidance.

\subsection{Explicit Theory-Driven Guidance}

As can be seen from Figure \ref{fig:main_fig}(b), drawing inspiration from Retrieval-Augmented Generation~\cite{rag} and Chain-of-Thought~\cite{cot} approaches, we retrieve dictionary information about the target word from the Oxford Dictionary\footnote{https://www.oxfordlearnersdictionaries.com} and derive multi-step thoughts from the LLM according to contents of metaphor theories. Sequentially, we explicitly prompt the LLM with the mentioned information to assist in making more reliable predictions.

In detail, given a sentence $s$ and a target word $w_t$, we first utilize NLTK\footnote{NLTK is a python library to process and analyze human language data.} to obtain the lemma of $w_t$. Next, we search for definitions and relevant usage examples of the lemma from the Oxford Dictionary. This collected information, denoted as $\mathcal{I}$, is then organized into a structured prompt that aids the LLM in developing a better semantic understanding of the target word with the context of the given sentence.

To stimulate the LLM's ability to apply metaphor theories, we generate the multi-step thoughts $\mathcal{X}$ by providing the LLM with the contents of MIP and SPV theories. The multi-step thoughts are also incorporated into the prompt to explicitly guide the LLM in making predictions based on these theories.

Ultimately, the gathered and generated information is provided to the LLM to obtain the response $R_{ex}$:

\begin{equation}
R_{ex}=\mathcal{F}(ins_{ex},\mathcal{X}, \mathcal{I}, s, w_t) ,
\label{eq:ex}
\end{equation}
where $ins_{ex}$ means the instruction for Explicit Theory-Driven Guidance.

\subsection{Self-Judgment}

Figure \ref{fig:main_fig}(c) illustrates the details of Self-Judgment. With the responses $R_{im}$ and $R_{ex}$, the final outputs are derived through a joint consideration of these two perspectives.

Specifically, responses $R_{im}$ and $R_{ex}$ offer rationales from different viewpoints, which may conflict with each other or be both incorrect. Therefore, we rely on the LLM to act as a judge, reviewing these responses and provide the final response $R_j$:

\begin{equation}
R_j=\mathcal{F}(ins_j, R_{im}, R_{ex}) ,
\label{eq:judge}
\end{equation}
where $ins_j$ means instruction for Self-Judgment. 

After the thorough verification of the responses from both perspectives, the ultimate answer is extracted from the response $R_j$.

\section{Experiments}

\subsection{Datasets}

\subfile{tabs/static}

We conduct experiments on two widely-used metaphor detection datasets: (1)\textbf{MOH-X}~\cite{moh-x} is sourced from WordNet~\cite{wordnet}, and contains sentences in which one verb is annotated as metaphorical or literal; (2)\textbf{TroFi}~\cite{TroFi} is another dataset dedicated to verb metaphor detection, including sentences from the 1987-89 Wall Street Journal Corpus Release 1. Following Tian et al.~\cite{TSI}, two balanced test sets are randomly sampled from these two datasets in our experiments.

Table \ref{tab:statistics} presents the statistics of these datasets. In the table, \textbf{\#Instances} indicates the total number of instances in dataset, \textbf{Ratio(\%)} reflects the percentage of instances containing metaphors, \textbf{Avg.L} denotes the average length of instances and \textbf{\#Samp} represents the number of randomly sampled balanced instances for evaluation.

\subsection{Implementation Details}

In our experiments, we utilize accuracy and F1-score as our evaluation metrics, both of which are reported the mean and standard deviation across 3 runs. Following Tian et al.~\cite{TSI},  both our framework and the baselines using LLMs are evaluated using OPENAI \textit{GPT-3.5 turbo}\footnote{gpt-3.5-turbo-0613} for a fair comparison. To explore the upper limits of our framework, we also assess its performance using \textit{GPT-4o}\footnote{gpt-4o-2024-08-06}, the high-intelligence flagship LLM of OPENAI.
The value of $k$ is set to 8 in Implicit Theory-Driven Guidance.\footnote{https://github.com/DeepLearnXMU/DMD}

\subfile{tabs/main_table}

\subsection{Main Results}

Table \ref{tab:main_res} presents the comparison results of DMD against various baselines, including methods without LLMs and methods with LLMs. The best and second-best results are in \textbf{bold} and \underline{underlined}, respectively.

The results indicate that our proposed DMD surpasses all baselines across all metrics on the two datasets. Compared to the best results among methods without LLMs, DMD yields a 6.32\% and 8.07\% increase in F1-score and a 6.00\% and 4.55\% boost in accuracy on the two datasets, respectively. This highlights that the explanations provided by DMD not only enhance the credibility of the predictions but also significantly boost the performance to metaphor detection. In comparison to TSI, the previous best-performing method with LLM,  our framework achieves a 3.47\% improvement in F1-score on the simpler MOH-X dataset and an even greater improvement of 4.36\% on the more challenging TroFi dataset. These results emphasize our framework’s superior capacity to deepen LLMs' metaphor comprehension and effectively tackle complex and ambiguous scenarios through rigorous reasoning.

Furthermore, the results of DMD on GPT-4o also reveal that DMD gains a significant performance enhancement, underscoring the DMD's broad applicability across different LLMs and its effective stimulation of LLMs' reasoning capabilities.

\subsection{Ablation Study}

Table \ref{tab:ablation}  presents the results from our ablation study on the MOH-X and TroFi datasets. To simplify the presentation of the table, we use the following abbreviations:\textit{ \textbf{Im}plicit Theory-Driven \textbf{G}uidance \textbf{(ImG)}, \textbf{Ex}plicit Theory-Driven \textbf{G}uidance \textbf{(ExG)}, }and\textit{ \textbf{S}elf-\textbf{J}udgment \textbf{(SJ)}.}

The results show that ExG outperforms ImG on the MOH-X dataset, highlighting the capability of the LLM to effectively utilize MIP and SPV theories for reasoning. In contrast, ImG exhibits superior performance on the more challenging TroFi dataset, indicating that the LLM is adept at leveraging theoretical feature similarities across samples. Importantly, ImG achieves substantial gains compared to vanilla few-shot method according to Table \ref{tab:main_res}, showcasing its ability to effectively extract theoretical insights from retrieved samples, leading to more accurate predictions. Furthermore, the results also point out the effectiveness of self-judgment in integrating both types of theory-driven guidance, with a significant improvement compared to using only ImG or only ExG.

Remarkably, each guidance of DMD is either comparable to or surpasses TSI, while avoiding the complexities associated with theoretical knowledge graph construction and multi-round question answering.

\subfile{tabs/ablation}

\section{Conclusion}

In this paper, we propose DMD, a framework that combines two perspectives, i.e., implicit and explicit, to guide LLMs on metaphor detection using metaphor theories including MIP and SPV. Within DMD, the Implicit Theory-Driven Guidance leverages metaphor theories to subtly influence the LLM's understanding of metaphors, whereas Explicit Theory-Driven Guidance directly prompt the LLM with well-designed instructions, providing the response from a different perspective. Finally, Self-Judgment synthesizes the viewpoints from both perspectives, leading to a more convincing response. Experimental results prove the effectiveness of DMD, with the SOTA performance on both datasets. We believe that our framework, with the novel integration of dual perspectives, offers a promising direction for future research in metaphor detection.

\bibliographystyle{IEEEtran}
\bibliography{refs}

\end{document}

%% file: imgs/main_fig.tex
\begin{figure*}[tbp]
\centering

    \includegraphics[width=\linewidth]{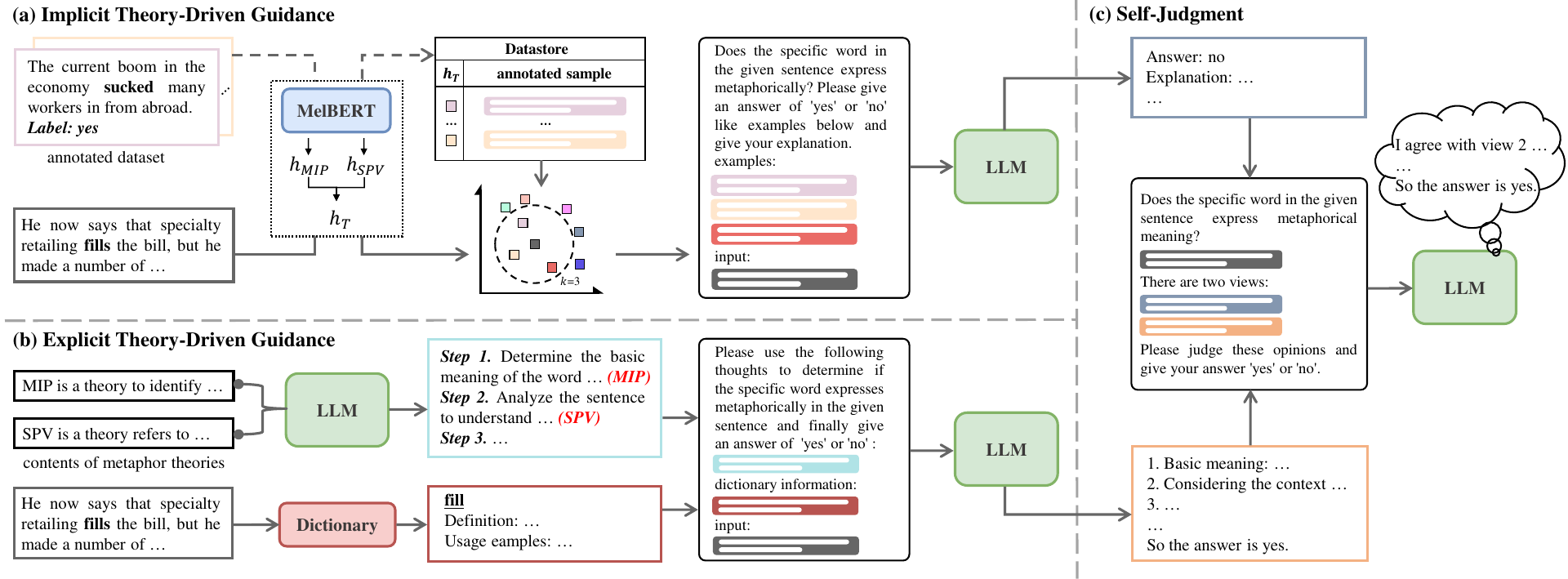}

\caption{Overview of our DMD framework. It consists of three parts: \textbf{(a) Implicit Theory-Driven Guidance:} For each sample, theoretical representations are computed using a pre-trained MelBERT model to find the $k$ nearest neighbors for in-context learning; \textbf{(b) Explicit Theory-Driven Guidance:} For each sample, the multi-step thoughts are generated by the LLM, and the information related to target word's lemma is retrieved from the Oxford Dictionary, both of which are used to guide the LLM explicitly; \textbf{(c) Self-Judgment:} Further evaluate the responses produced by Implicitly Theory-Driven Guidance and Explicitly Theory-Driven Guidance, and the final answer is extracted from the judgment.}

\label{fig:main_fig}
\end{figure*}

%% file: tabs/static.tex
\begin{table}[]

\centering

\caption{Statistics of the datasets}

\label{tab:statistics}

\begin{tabular}{ccccc}
\toprule
\textbf{Dataset} & \textbf{\#Instances} & \textbf{Ratio(\%)} & \textbf{Avg.L} & \textbf{\#Samp} \\ \midrule
MOH-X & 647 & 48.69 & 8.0 & 300 \\
TroFi & 3737 & 43.54 & 28.3 & 300 \\ \bottomrule
\end{tabular}

\end{table}

%% file: tabs/main_table.tex
\begin{table}[tbp]
\centering

\caption{Main results}
\label{tab:main_res}

\resizebox{\linewidth}{!}{
\begin{tabular}{lccccr}
\toprule
\multicolumn{1}{c}{}               & \multicolumn{2}{c}{\textbf{MOH-X}} & \multicolumn{2}{c}{\textbf{TroFi}} \\ \cmidrule(l){2-3} \cmidrule(l){4-5} 
\multicolumn{1}{l}{\multirow{-2}{*}{\textbf{Method}}} & F1       & Acc.    & F1       & Acc.    \\ \midrule
\multicolumn{5}{l}{\cellcolor[HTML]{EFEFEF}\textit{\textbf{Methods without LLM}}}                \\
MelBERT\cite{melbert-tra3}  & 77.88 ± 0.83   & 77.89 ± 0.83  & 62.36 ± 1.51   & 62.89 ± 1.29  \\
MisNet\cite{misnet}   & 77.08 ± 1.12   & 77.11 ± 1.13  & 62.01 ± 0.64   & 62.67 ± 0.54  \\
AdMul\cite{admul}    & 79.74 ± 0.44   & 79.89 ± 0.42  & 60.54 ± 1.43   & 62.67 ± 0.98  \\
\multicolumn{5}{l}{\cellcolor[HTML]{EFEFEF}\textit{\textbf{Methods with LLM}}}                  \\
Zero-shot\cite{zero-shot}   & 66.43 ± 1.45      & 69.11 ± 1.29     & 58.58 ± 1.01      & 61.22 ± 0.95     \\
Zero-shot CoT\cite{cot}        & 70.97 ± 0.37      & 71.22 ± 0.32     & 64.18 ± 1.07      & 64.78 ± 0.95     \\
Few-shot\cite{few-shot}            & 76.78 ± 2.95      & 75.55 ± 3.46     & 61.53 ± 1.98      & 52.11 ± 1.68
  \\
Plan-and-Solve\cite{plan_and}       & 54.27 ± 2.73      & 57.89 ± 2.01     & 58.13 ± 2.29      & 58.33 ± 2.23     \\
Self-refine\cite{self-ref}          & 64.06 ± 0.37      & 67.67 ± 0.27     & 57.27 ± 0.22      & 60.89 ± 0.16     \\
Self-consistency\cite{self-consis}     & 70.65 ± 1.58      & 72.22 ± 1.50     & 60.30 ± 0.76      & 61.56 ± 0.68     \\
Least-to-most\cite{least2most}        & 74.43 ± 1.27      & 75.00 ± 1.25     & 63.88 ± 3.16      & 64.11 ± 3.03     \\
TSI\cite{TSI}                  & 82.59 ± 2.22      & 82.93 ± 1.94     & 66.07 ± 1.11      & 66.89 ± 1.13     \\
\textbf{DMD (GPT-3.5 turbo)}   & \underline{86.06 ± 1.61}   & \underline{85.89 ± 1.64}  & \underline{70.43 ± 0.98}   & \underline{67.44 ± 1.01}  \\ 
\textbf{DMD (GPT-4o)}   & \textbf{91.71 ± 0.95}   & \textbf{91.53 ± 1.07}  & \textbf{73.73 ± 0.53}   & \textbf{70.78 ± 0.77}  \\   \bottomrule
\end{tabular}
}
\vspace{-5pt}
\end{table}

%% file: tabs/ablation.tex
\begin{table}[t]

\caption{Ablation experimental results}
\label{tab:ablation}

\resizebox{\linewidth}{!}{
\begin{tabular}{ccccccc}
\toprule
\multirow{2}{*}{\textbf{ImG}} & \multirow{2}{*}{\textbf{ExG}} & \multirow{2}{*}{\textbf{SJ}} & \multicolumn{2}{c}{\textbf{MOH-X}} & \multicolumn{2}{c}{\textbf{TroFi}} \\ \cmidrule(l){4-5} \cmidrule(l){6-7} 
 &  &  & F1 & Acc. & F1 & Acc. \\ \midrule
\ding{55} & \ding{55} & \ding{55} & 66.43 ± 1.45 & 69.11 ± 1.29 & 58.58 ± 1.01 & 61.22 ± 0.95 \\
\ding{51} & \ding{55} & \ding{55} & 82.85 ± 0.66 & 81.55 ± 0.77 & 68.59 ± 1.01 & 66.22 ± 0.69 \\
\ding{55} & \ding{51} & \ding{55} & 83.53 ± 1.01 & 82.76 ± 1.78 & 66.36 ± 0.11 & 65.56 ± 0.19 \\
\ding{51} & \ding{51} & \ding{51} & 86.06 ± 1.61 & 85.89 ± 1.64 & 70.43 ± 0.98 & 67.44 ± 1.01 \\ \bottomrule
\end{tabular}
}
\vspace{-5pt}

\end{table}